\documentclass{article}

\usepackage{PRIMEarxiv}

\usepackage[utf8]{inputenc} 
\usepackage[T1]{fontenc}    
\usepackage{hyperref}       
\usepackage{url}            
\usepackage{booktabs}       
\usepackage{amsfonts}       
\usepackage{nicefrac}       
\usepackage{microtype}      
\usepackage{lipsum}
\usepackage{fancyhdr}       
\usepackage{graphicx}       
\graphicspath{{media/}}     
\usepackage{multirow}

\title{Semi-MAE: Masked Autoencoders for Semi-supervised Vision Transformers
}

\author{
  Haojie Yu, Kang Zhao, Xiaoming Xu \\
  Meituan Inc. \\
  \texttt{\{yuhaojie02, zhaokang, xuxiaoming04\}@meituan.com} \\
}

\begin{document}
\maketitle

\begin{abstract}
Vision Transformer (ViT) suffers from data scarcity in semi-supervised learning (SSL). To alleviate this issue, inspired by masked autoencoder (MAE), which is a data-efficient self-supervised learner, we propose Semi-MAE, a pure ViT-based SSL framework consisting of a parallel MAE branch to assist the visual representation learning and make the pseudo labels more accurate. The MAE branch is designed as an asymmetric architecture consisting of a lightweight decoder and a shared-weights encoder. We feed the weakly-augmented unlabeled data with a high masking ratio to the MAE branch and reconstruct the missing pixels. Semi-MAE achieves 75.9\% top-1 accuracy on ImageNet with 10\% labels, surpassing prior state-of-the-art in semi-supervised image classification. In addition, extensive experiments demonstrate that Semi-MAE can be readily used for other ViT models and masked image modeling methods.
\end{abstract}


\section{Introduction}
To date, Vision Transformers (ViT)\cite{dosovitskiy2020image} have achieved significant progress in supervised learning\cite{dosovitskiy2020image,touvron2021training,liu2021swin}, self-supervised learning\cite{chen2020generative,Chen_2021_ICCV,xie2022simmim,bao2021beit,he2022masked}, and various other computer vision tasks\cite{carion2020end,zhu2020deformable,radford2021learning,ramesh2021zero}. ViTs have a weaker inductive bias than CNNs\cite{dosovitskiy2020image}, therefore a large amount of training data is often required to make ViTs generalize well. As a consequence, the performance of ViTs is unsatisfactory in semi-supervised learning (SSL), where only a small number of labeled data is provided and the rest are unlabeled. As discussed in \cite{weng2022semi}, using FixMatch\cite{sohn2020fixmatch}, one of the most popular SSL approaches, to train a ViT presents an inferior performance than CNN architectures. To tackle this problem, \cite{weng2022semi} proposes a joint semi-supervised training of CNN and ViT that outperforms CNN counterparts. Further, we continue to explore pure ViTs in SSL, intending to obtain more accurate pseudo labels by improving the visual representation of ViT itself.

In this work, we propose a simple yet efficient SSL paradigm of pure ViTs that surpasses previous CNN-based methods. We consider the success of Transformers\cite{vaswani2017attention} in self-supervised learning\cite{dosovitskiy2020image}, which is a promising solution to the data scarcity issue by leveraging a large amount of unlabeled data. Specifically, we build on FixMatch\cite{sohn2020fixmatch} framework except replace all CNNs with ViTs. Then for enhancing the learning of visual representations, we introduce a masked autoencoder (MAE)\cite{he2022masked} branch, which is parallel to the SSL framework and they share the same encoder. We mask random patches from unlabeled data and design a lightweight decoder to reconstruct the input. Accordingly, the mean squared error (MSE) between the reconstructed and original images contributes to the final loss. We call our method Semi-MAE. Note that Semi-MAE is scalable to any other transformer-based model.


We perform extensive experiments to evaluate Semi-MAE. Notably, Semi-MAE with ViT-Small reaches 75.9\% top-1 accuracy on ImageNet with 10\% labeled images. We demonstrate that pure ViTs can outperform CNN-based\cite{sohn2020fixmatch,zhai2019s4l,Pham_2021_CVPR} and joint\cite{weng2022semi} SSL frameworks. Additionally, our MAE branch is a plug-and-play module that can improve Semiformer\cite{weng2022semi} by 0.9\%. We also show that other masked image modeling methods can further bring gains for Semi-MAE, e.g., 76.0\% top-1 accuracy with LoMaR\cite{chen2022efficient}.

    
    

\section{Related Work}
\paragraph{Vision Transformers}
Transformers\cite{vaswani2017attention} have made substantial achievements in natural language processing (NLP)\cite{devlin2018bert}, but in computer vision, convolutional neural networks (CNN) have dominated the past decade due to their image-specific inductive bias. The appearance of \cite{dosovitskiy2020image}, Vision Transformers (ViT) have finally addressed the architectural gap and have achieved success in image recognition\cite{dosovitskiy2020image,yuan2021tokens,liu2021swin}, object detection\cite{carion2020end,zhu2020deformable}, segmentation\cite{wang2021max,zheng2021rethinking}, etc. However, ViTs have encountered obstacles when applied to semi-supervised learning (SSL), where the amount of labeled data is insufficient for its training. In this work, we provide Semi-MAE to solve the aforementioned challenge.

\paragraph{Masked image modeling}
Breakthroughs in masked language modeling (MLM)\cite{devlin2018bert} in NLP have generated great interest in the computer vision community, leading to the birth of masked image modeling (MIM) methods. \cite{dosovitskiy2020image} studied the masked patch prediction objective and surprisingly found that self-training worked quite well on few-shot metrics. iGPT\cite{chen2020generative} trained a sequence transformer to auto-regressively predict pixels on low-resolution ImageNet. BEiT\cite{bao2021beit} first tokenized image patches into visual tokens via discrete VAE and then predicted randomly masked visual tokens by the corrupted original image patches. Recently, masked autoencoder (MAE)\cite{he2022masked} proposed an autoencoding approach, whose objective was simply to reconstruct missing original patches in the pixel space given a partial observation. The asymmetric design and high masking ratio yield a nontrivial task and help to learn well-generalized models while leading to a significant reduction in computation.

\paragraph{Semi-supervised learning}
Semi-supervised learning (SSL) has been shown to be a promising solution to exploit unlabeled data. There are two classic strategies for SSL. One is pseudo labeling\cite{lee2013pseudo,rosenberg2005semi} where model predictions are converted to hard labels, the other is consistency regularization\cite{bachman2014learning,sajjadi2016regularization} where models are trained to output consistent results for different views of the input. FixMatch\cite{sohn2020fixmatch} integrated these two strategies: on unlabeled data, hard pseudo labels are generated with weak augmentation as the target, and the model is fed a strongly-augmented version of the same image. Motivated by ViTs' success, \cite{weng2022semi} proposed a joint semi-supervised training of CNN and ViT. For the first time, the application of ViTs in SSL achieves comparable performance against the CNN counterparts. On top of this, we continue to explore the pure ViTs in SSL.

\begin{figure}
  \centering
  \includegraphics[width=1.0\textwidth]{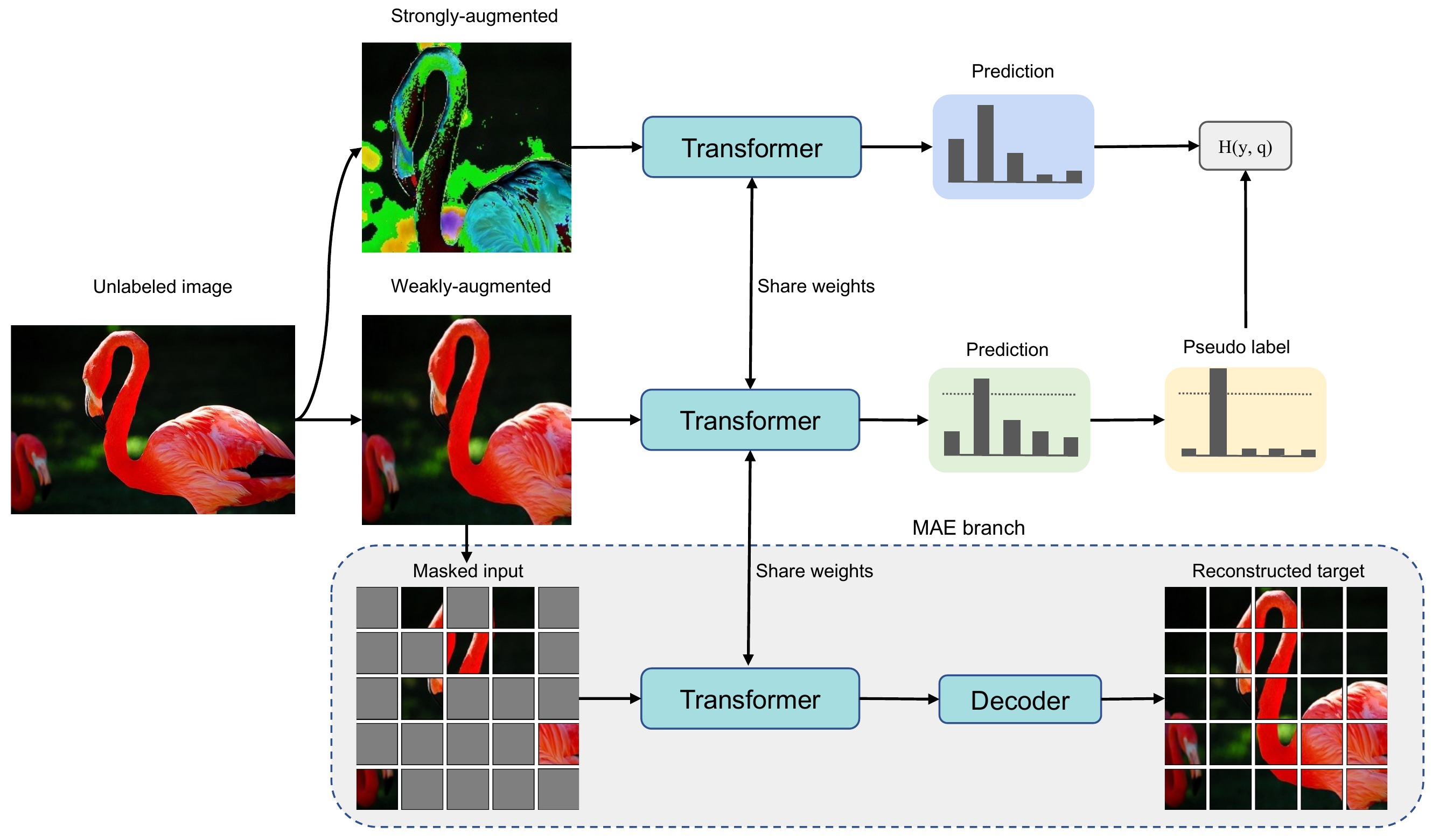}
  \caption{Overview of Semi-MAE. A weakly-augmented unlabeled sample is fed into the ViT, whose predictions with a high confidence score can be converted to a one-hot pseudo label. The pseudo label is used for supervising the model's predictions of the strongly-augmented version of the same unlabeled sample. At the same time, the weakly-augmented sample is divided into image patches. A random subset of patches is masked out and the rest are taken as the input of the MAE branch. This branch shares the same ViT as the SSL framework. The latent representation and mask tokens are processed by a small decoder to reconstruct the original image.}
  \label{fig:model}
\end{figure}

\section{Semi-MAE}
\label{sec:semimae}
Semi-MAE is a semi-supervised learning (SSL) framework of pure Vision Transformers\cite{dosovitskiy2020image} (ViTs) as illustrated in Figure \ref{fig:model}. Building on FixMatch\cite{sohn2020fixmatch}, Semi-MAE introduces a masked autoencoder (MAE) branch to assist the encoder's visual representation learning. This branch parallels the original SSL framework and shares encoder weights. In this section, we will first review FixMatch\cite{sohn2020fixmatch} algorithm and then elaborate proposed MAE branch. 

\paragraph{Base architecture}
FixMatch\cite{sohn2020fixmatch} is one of the most popular SSL frameworks in recent times. Its main contribution lies in the combination of consistency regularization and pseudo-labeling. Specifically, the overall loss consists of two cross-entropy losses: a supervised loss $L_s$ and an unsupervised loss $L_u$. For a labeled sample $\{(x^l_i, y^l_i)\}^{N_l}_{i=1}$, a weak augmentation $\alpha(\cdot)$ is applied to compute the supervised loss
\begin{equation}
    L_s = \frac{1}{N_l} \sum^{N_l}_{i=1}H(y_i, f(\alpha(x^l_i))),
\end{equation}
where $H(\cdot, \cdot)$ and $f(\cdot)$ are respectively the cross entropy loss function and the model forward function. As for an unlabeled sample $\{x^u_i\}^{N_u}_{i=1}$, we first compute the output probabilities of its weakly-augmented version $p_i = f(\alpha(x^u_i))$. The pseudo label is produced by $\hat{y}_i = argmax(p_i)$ with its confidence $max(p_i)$. The unsupervised loss is calculated on the strong-augmented sample
\begin{equation}
    L_u = \frac{1}{N_u} \sum^{N_u}_{i=1}H(\hat{y}_i, f(\mathcal{A}(x^u_i)))\delta(max(p_i) > \tau),
\end{equation}
where $\mathcal{A}(\cdot)$ and $\delta(\cdot)$ are respectively the strong augmentation and the indicator function. $\tau$ is the confidence threshold. The overall loss is just $L = L_s + \lambda L_u$ where $\lambda$ denotes the relative weight of the unsupervised loss. In Semi-MAE, we simply replace CNN models with ViTs.

\paragraph{MAE branch}
As figured out in \cite{weng2022semi}, the performance of ViT in SSL building on FixMatch\cite{sohn2020fixmatch} is inferior to CNN counterparts. ViT generates inaccurate pseudo labels due to limited labeled data. Therefore we introduce a masked autoencoder (MAE)\cite{he2022masked} branch to enforce the visual representation learning and help ViT generate more accurate pseudo labels. Masked autoencoder learns visual representations from images with a high masking ratio and has demonstrated its efficiency and effectiveness even only on ImageNet\cite{deng2009imagenet}. In particular, weakly-augmented unlabeled data also serve as the original input of the MAE branch. The original input is first divided into patches, then we mask a high proportion of patches and feed the rest to our encoder. This encoder is a ViT that shares the same weights as the encoder in the SSL framework. Following the asymmetric design in \cite{he2022masked}, a small and independent decoder is used to reconstruct the corrupted image from the latent representation and mask tokens. The reconstructed target is the pixel value for each masked patch. The loss function is the mean square error (MSE) between the reconstructed and original images in the pixel space. Eventually, our total loss function is
\begin{equation}
    L = L_s + \lambda L_u + \mu L_{MAE}
\end{equation}
where $\mu$ is the trade-off for self-supervised loss weight.

\section{Experiments}
\subsection{Experiment Settings}
\paragraph{Datasets and evaluation metric}
We conduct experiments on ImageNet\cite{deng2009imagenet}, which contains $\sim$1.28M training and 50K validation images. Following \cite{sohn2020fixmatch}, we sample 10\% labeled images from the ImageNet training set and leave the rest as unlabeled data. We select top-1 accuracy on the validation set as the evaluation metric. In addition, for a fair comparison, we apply the same data augmentation as \cite{weng2022semi}.

\paragraph{Implementation details}
We train the model from scratch. In detail, we first warm up the model for 100 epochs and then train the model for 600 epochs with semi-supervision. We apply AdamW\cite{loshchilov2017decoupled} as the optimizer with an initial learning rate $10^{-3}$, which decays towards $10^{-5}$ using the cosine decay scheduler. The trade-offs $\lambda$ and $\mu$ are respectively 10.0 and 5.0. In each batch, the ratio between labeled and unlabeled images is 1:7. We mainly use ViT-Small\cite{dosovitskiy2020image} as our backbone. As for the MAE branch, we follow the default settings of \cite{he2022masked}.

\subsection{Main Results}
We compare Semi-MAE with the state-of-the-art semi-supervised methods. Results are presented in Table \ref{tab:main results}. Using only ViT-Small, which has a smaller number of parameters than ResNet-50(22M v.s. 26M), Semi-MAE achieves 75.9\% top-1 accuracy that outperforms the prior state-of-the-art CNN-based methods. Semiformer\cite{weng2022semi} first introduces ViT to SSL and achieves 75.5\% top-1 accuracy with a joint framework. However, Semi-MAE with ViT-S alone can further improve the performance over 0.4\% than Semiformer\cite{weng2022semi}. These comparisons demonstrate that Semi-MAE achieves state-of-the-art performance without additional data and more architectural improvement.

\begin{table}[htbp]
 \caption{The comparisons with state-of-the-art models.}
  \centering
  \begin{tabular}{lcc}
    \toprule
    Method     & Architecture     & Top-1 Acc(\%) \\
    \midrule
    UDA\cite{xie2020unsupervised} & ResNet-50 & 68.8\% \\
    FixMatch\cite{sohn2020fixmatch} & ResNet-50 & 71.5\% \\
    S4L\cite{zhai2019s4l} & ResNet-50(4x) & 73.2\% \\
    MPL\cite{Pham_2021_CVPR} & ResNet-50 & 73.9\% \\
    CowMix\cite{french2020milking} & ResNet-50 & 73.9\% \\
    Semiformer\cite{weng2022semi} & ViT-S+ResNet50 & 75.5\% \\
    \hline
    Semi-MAE (ours) & ViT-S & 75.9\% \\
    \bottomrule
  \end{tabular}
  \label{tab:main results}
\end{table}

\subsection{Ablation Studies}
\label{sec:ablation}
\paragraph{MAE branch}
To prove the effectiveness and efficiency of our MAE branch, we further implement it into Semiformer\cite{weng2022semi}. Results in Table \ref{tab:mae branch} present that the MAE branch can bring marginal gains of 0.9\% over the baseline.

\begin{table}[h!]
 \caption{Results of Semiformer\cite{weng2022semi} with MAE branch.}
  \centering
  \begin{tabular}{lccc}
    \toprule
    Method & Architecture & MAE branch & Top-1 Acc(\%) \\
    \midrule
    \multirow{2}{*}{Semiformer} & \multirow{2}{*}{ViT-S+ResNet50} & & 75.5\% \\
     & & \checkmark & 76.4\% \\
    \bottomrule
  \end{tabular}
  \label{tab:mae branch}
\end{table}

\paragraph{Other masked image modeling methods}
Several masked image modeling methods\cite{chen2022efficient,bao2021beit,he2022masked} have demonstrated their effectiveness to learn visual representations from images. Therefore, we investigate other MIM methods besides MAE\cite{he2022masked} and observe that LoMaR\cite{chen2022efficient} can further boost the model performance by 0.1\%, as shown in Table \ref{tab:MIM methods}.

\begin{table}[h!]
 \caption{Results with other masked image modeling methods.}
  \centering
  \begin{tabular}{lcc}
    \toprule
    MIM Method     & Architecture     & Top-1 Acc(\%) \\
    \midrule
    MAE & ViT-S & 75.9\%  \\
    LoMaR & ViT-S & 76.0\% \\
    \bottomrule
  \end{tabular}
  \label{tab:MIM methods}
\end{table}

\section{Conclusion}
We propose Semi-MAE, a pure Vision Transformer-based semi-supervised learning framework. By introducing a masked autoencoder branch, Semi-MAE achieves substantial performance without extra training data. On ImageNet with 10\% labels, Semi-MAE can reach 75.9\% top-1 accuracy, which surpasses the state-of-the-art CNN-based and joint semi-supervised methods. This has proven that pure Vision Transformer is a promising solution for semi-supervised learning.

\bibliographystyle{unsrt}  
\bibliography{references}

\end{document}